# Leveraging Reinforcement Learning Techniques for Effective Policy Adoption and Validation

Nikki Lijing Kuang[1] and Clement H. C. Leung[2]

[1] Department of Computer Science and Engineering
University of California, San Diego, La Jolla, CA, USA
`l1kuang@ucsd.edu`
[2] School of Science and Engineering
Chinese University of Hong Kong, Shenzhen
`clementleung@cuhk.edu.cn`

**Abstract.** Rewards and punishments in different forms are pervasive and present in a wide variety of decision-making scenarios. By observing the outcome of a sufficient number of repeated trials, one would gradually learn the value and usefulness of a particular policy or strategy. However, in a given environment, the outcomes resulting from different trials are subject to chance influence and variations. In learning about the usefulness of a given policy, significant costs are involved in systematically undertaking the sequential trials; therefore, in most learning episodes, one would wish to keep the cost within bounds by adopting learning stopping rules. In this paper, we examine the deployment of different stopping strategies in given learning environments which vary from highly stringent for mission critical operations to highly tolerant for non-mission critical operations, and emphasis is placed on the former with particular application to aviation safety. In policy evaluation, two sequential phases of learning are identified, and we describe the outcomes variations using a probabilistic model, with closed-form expressions obtained for the key measures of performance. Decision rules that map the trial observations to policy choices are also formulated. In addition, simulation experiments are performed, which corroborate the validity of the theoretical results.

## 1  Introduction

In order to determine the feasibility or optimality of a given course of action, it is necessary to observe and monitor the outcomes of repeated trials. Repetition is necessary in order to ensure reliability and ongoing effectiveness particularly in an environment which is subject to chance influence and where complete information on all the underlying factors are not available. This is especially vital for mission critical operations, such as aviation safety, where ongoing validation or monitoring of a given policy is essential, but is also relevant for non-mission critical activities [18][19][21]. In most practical situations, such trials cannot be performed in parallel but have to be undertaken in a sequential manner. In the context of reinforcement learning, each outcome can be classified as a reward or punishment. However, the cost of carrying out such learning trials can be significant and such sequential validation can have varying degrees of stringency. In this study, the stochastic structure of the environment is explicitly modeled, and the performance measures of the associated validation costs are analyzed.

In trialing or learning a given course of action, the observed rewards and punishments are usually probabilistic. For instance, when one is experimenting a new route between an originating point A and a destination B, an increase of the journey duration by a given amount may be viewed as a punishment, whereas a reduction in the duration of the same may be viewed as a reward, and having thus learned, say, the acceptability of the new route, one would adopt the new route as a policy for traveling between A and B.

Thus, through repeated trials resulting in outcomes of either reward or punishment, one establishes the feasibility of the new policy and completes the learning phase. Subsequent to the learning phase, the new policy, if learned successfully (i.e. when the rewards to punishments ratio is sufficiently high) is adopted from that point onwards without it being questioned or evaluated afterwards. In this particular example, the learning is primarily done during the pre-adoption phase. In some situations, however, even after the policy is adopted, ongoing validation and monitoring is still carried out and this is especially necessary for safety-critical and mission-critical operations. If in the course of ongoing monitoring, there is an overwhelming number of punishments observed, then the adoption of the policy may be called into question, and termination of the policy may be necessary.

In this paper, we study such learning reinforcement scenarios of stochastically receiving rewards and punishments for both the pre-adoption phase as well as the post-adoption phase. To be concrete, we shall use a scenario of aviation safety, as we believe this scenario is sufficiently general and of particular relevance and currency to present day concerns. Despite this, we wish to point out that many other everyday learning situations are similar to this; examples include trialing a new machine translation algorithm, learning the effectiveness of a new advertising channel, and route discovery in self-drive vehicles.

An autonomous agent in reinforcement learning learns through the interaction with the environment to maximize its rewards, while minimizing or avoiding punishments. In most practical situations, the underlying environment is non-stationary and noisy [1][4][6][20][22], and the next state results from taking the same policy may not result in the same outcome every time but appears to be stochastic [2][7]. In [3]. Brafman and Tennenholtz introduces a model-based reinforcement learning algorithm R-Max to deal with stochastic games [5]. Such stochastic elements can notably increase the complexity in multi-agent systems and multi-agent tasks, where agents learn to cooperate and compete simultaneously [6][10]. As other agents adapt and actively adjust their policies, the best policy for each agent would evolve dynamically, giving rise to non-stationarity [8][9]. In these studies, the cost of a trial to receive either a reward or punishment can be seen to be significant, and ideally, one would like to arrive at the correct conclusion by incurring minimum cost. In reinforce learning algorithms, we are always in the hope to rapidly converge to an optimal policy with least volumes of data, calculations, learning iterations, and minimal degree of complexity [11][12]. To do so, one should explicitly define the stopping rules for specifying the conditions under which learning should terminate and a conclusion drawn as to whether the learning has been successful or not based on the observations so far.

Establishing stopping rules, is an active research topic in reinforcement learning, which is closely linked to the problems of optimal policies and policy convergence [13]. Conventional approaches mainly aim for relatively small-scale problems with finite states and actions. The stopping rules involved are well-defined for each category of algorithms, such as utilizing Bellman Equation in $Q$-learning [14]. To deal with continuous action spaces or state spaces, new algorithms, such as the Cacla algorithm [15] and CMA-ES algorithm [16], are developed with specific stopping rules. Some stopping rules for stochastic reinforcement learning under different assumptions have been proposed and studied in [23][24]. Still, most studies on stopping rules are procedure-oriented and do not have a unified measurement where comparison may be facilitated.

In our study here, in addition to learning from the observations in the pre-adoption and post-adoption phases, we also focus in the stopping criteria, so that what has been observed and learned can form the basis of policy decision making. The next section provides a representation of the stochastic learning environment, and establishes stopping rules for the different phases. An analysis of these rules is given in Section III. Assessment of the learning cost and the rewards ratio, along with experimental evaluation is given in Section IV, and the final conclusions are drawn in Section V.

## 2     The Learning Environment and Stopping Rules

We assume that trials are systematically carried out in a sequential manner. Due to the presence of a multiplicity unknown factors and hidden variables, indicating an environment about which we have incomplete information, the outcome from different trials will be subject to probabilistic influences. As mentioned earlier, we shall employ the aviation safety learning situations to develop the main ideas. The reason for using this situation is twofold:
   i.   it has a high degree of generality that is able to subsume a variety of learning situations as special cases, and
   ii.  it has a particular relevance and interest to current concerns of airlines, aircraft manufacturers, and passengers.

We shall divide the learning reinforcement of a policy into two distinct phases:
   i.   the pre-adoption phase, which we shall refer to as Phase I, and
   ii.  the post-adoption phase, which we shall refer to as Phase II.

The former phase is concerned with learning the acceptability of the policy through systematic trials, while the latter is concerned with the continued validity of the policy subsequent to adoption, and in this case, whether the policy should under some circumstances be discontinued. An especially relevant example is whether a particular aircraft model recently introduced should continue to be in service or should it be discontinued, at least temporarily, for the safety and well-being of its passengers, perhaps following some serious incidents.

Here, we are dealing with a sequence of independent learning trials, each of which either results in a reward or punishment. In our particular aviation example, typically for each trial a set of indicators are logged and a final score is computed which forms the basis of a decision on either a pass (reward) or failure (punishment) for the trial is attained. We let $p$ and $q$, with $p + q = 1$, denote the probabilities of receiving a reward or punishment respectively for a given trial. For example, if $p >> q$, then the decision should be that of adopting the policy. In general the requirements for Phase I is much more stringent for Phase II, and the cost of different decision rules will be analyzed in the next section.

Let us consider the following two stopping rules.

***Rule I****: In the course of undertaking the learning trials, an agent concludes the learning process when m consecutive rewards are obtained.*

***Rule II****: In the course of undertaking the learning trials, an agent concludes the learning process when m total rewards are obtained.*

Rule I is a somewhat stringent stopping rule but is particular applicable for mission critical operations where a high degree of reliability is required. It is also more widely used for the proper learning phase (Phase I) than for the validation phase after learning (Phase II). Rule II is a less stringent stopping rule and is often used for the validation phase. In some applications, such as finding an optimal route from A to B for a self-driving vehicle, it is mostly sufficient just to use Rule II for Phase I learning, and usually no need for Phase II evaluation.

In addition, there is a significant difference between the objective of Phase I, and that of Phase II. While the objective of Phase I is to aim to adopt the policy by accumulating a sufficient number of rewards, the aim of Phase II, on the other hand, is to look for alerts that may lead to a discontinuation of the policy. As we shall see, the analysis in Phase II requires the application of the Reflection Principle [17], by interchanging the probabilities $p$ and $q$, as well as interchanging the rewards and punishments. Such reversal of roles leads to a variation of Rule I and Rule II, which we shall call Rule IR, and Rule IIR respectively, with the suffix R signifying reflection.

***Rule IR****: In the course of undertaking the validation trials, an agent concludes the learning process when m' consecutive punishments are received.*

***Rule IIR****: In the course of undertaking the validation trials, an agent concludes the learning process when m' total punishments are received.*

As we shall see in the next section, the use of Rule I for Phase I means that acceptance of the policy is more stringent than that when Rule II is used. On the other hand, the use of IIR in Phase II also signifies a more stringent requirement since rejection of the policy is easier than that of using Rule IR.

**Table 1.** The typical learning scenarios for different types of applications for the two phases.

|  | PHASE I (LEARNING) | PHASE II (VALIDATION) |
|---|---|---|
| **MISSION CRITICAL SYSTEMS** | Rule I | Rule IIR |
| **INTERMEDIATE LEVEL 2** | Rule I | Rule IR |
| **INTERMEDIATE LEVEL 1** | Rule II | Rule IIR |
| **NON-MISSION CRITICAL SYSTEMS** | Rule II | IR |

## 3  Analysis of the Performance of the Stopping Rules

Rule I above is concerned with collecting a given number of consecutive reinforcements or rewards, so that we shall first establish the probability of occurrence of such an event for the first time. Let bn be the probability that m consecutive rewards occurs at trial n, with n ≥ m, not necessarily for the first time, and we denote by *B(z)* be the corresponding probability generating function. From [17], this probability generating function can be obtained as

$$B(z) = \frac{1 - z + qp^m z^{m+1}}{(1-z)(1-p^m z^m)}. \tag{1}$$

Since we need to obtain the corresponding generating function for the probability that the associated event occurs for the first time, we need to consider the relationship between the two events. We shall use the random variable *X* to denote the number of trials preceding and including the receiving of the first set of *m* consecutive rewards. Thus *X* is the stopping time for Rule I, measured in terms of the number of trials, and we let $a_n$ be the probability

$$a_n = \Pr[X = n], \quad n = m, m+1, \dots. \tag{2}$$

We denote by *A(z)* the probability generating function for the event that the accumulation of *m* rewards occurs for the first time. It can be shown in [17] that the generating function *A(z)* is related to *B(z)* by

$$A(z) = \frac{B(z) - 1}{B(z)}. \tag{3}$$

From this, we obtain, after simplification,

$$A(z) = \frac{p^m z^m}{1 - q^m \sum_{k=0}^{m-1} p^k z^k}. \tag{4}$$

From this, the mean and variance of *X* can be readily obtained after simplification,

$$\mathrm{E}[X] = A'^{(1)} = \frac{1-p^m}{qp^m}, \tag{5}$$

$$\mathrm{Var}[X] = A''(1) + A'(1) - A'(1)^2 = \frac{1}{q^2 p^{2m}} - \frac{2m+1}{qp^m} - \frac{p}{q^2}, \tag{6}$$

where the apostrophe indicates derivative.

Next, we examine Rule II, and let the random variable $Y$ be the number of observations preceding and including the first reward; thus

$$\Pr[Y = k] = pq^{k-1}, \quad k = 0, 1, 2, 3, \ldots \tag{7}$$

The probability generating function for a random variable $W$ which excludes the reward itself has been obtained in [23] and is given by

$$\frac{p}{(1-qz)}.$$

The random variables $W$ and $Y$ are related by $Y = W + 1$, and since the generating function of 1 is $z$, we have, for the probability generating function of $Y$,

$$\frac{pz}{(1-qz)}. \tag{8}$$

Since after the occurrence of the first reward, the process probabilistically repeats itself again, so that we have for the number of trials to the $m^{\text{th}}$ reward, bearing in mind that under Rule II, the rewards need not occur consecutively

$$Z = \sum_{k=1}^{m} Y_k, \tag{9}$$

where each $Y_k$ has the same distributional characteristics as $Y$. Consequently, the probability generating function of $F(z)$ corresponding to $Z$ may be obtained

$$F(z) = \left[\frac{pz}{(1-qz)}\right]^m \tag{10}$$

From this, the mean and variance of $Z$ can be readily obtained by differentiation,

$$\mathrm{E}[Z] = \frac{m}{p}, \tag{11}$$

$$\mathrm{Var}[Z] = \frac{mq}{p^2}. \tag{12}$$

It is not hard to see that $\mathrm{E}(X) \geq \mathrm{E}(Z)$, since achieving $m$ consecutive rewards necessarily implies achieving at least $m$ total non-consecutive rewards, with equality holding iff $m=1$, since in this case, there is no difference between the two situations.

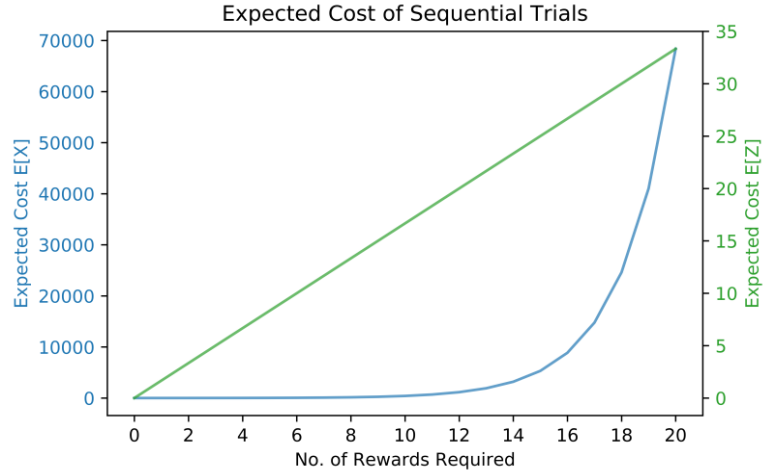

**Fig. 1.** Cost Comparison of Rules I and II ($p = 0.6$).

Figure 1 compares the average cost of sequential trials of the two rules. Here, the left vertical axis is used for E(**X**)**,** while, the right vertical axis is used for E(**Z**)**.** We see that the stringency of Rule I is manifested in a steep climb in the number of trials as *m* increases, as opposed to a relatively moderate increase in Rule II.

## 4      Learning Cost Evaluation and Experimentation

The number of trials carried out to complete the learning episode is often a costly process. Let *h* be the numerical representation of cost associated with a single trial, and we can standardize on *h* as the cost unit so that and without loss of generality we can set *h* = 1. Having specified *m*, a minimum observation cost of *mh* must therefore be incurred. What is uncertain is the number of punishments obtained in the process, and ideally to achieve minimum cost, this number should be small. Such average trialing cost are given by Equations (5) and (11). Simulation experiments are carried out to compare the actual average learning cost with theoretical predictions, and these are given in Table 2. Table 3 gives the corresponding comparisons of the standard deviation from Equations (6) and (12).

The mean number of trials required in order to accumulate *m* rewards has a direct bearing on the adoption of the given policy. The learning overhead, or cost ratio, is given by the ratio of the average total number of trials to the number of required rewards *m*. For Rule I, this is given by

$$r_1(m) = \frac{1 - p^m}{mqp^m},  \qquad(13)$$

and for Rule II, this is given by,

$$r_2(m) = \frac{1}{p}. \tag{14}$$

Clearly, the decision for adoption or successful validation will tend to be positive for small $r_1$ and $r_2$, but tends towards negative when $r_1$ and $r_2$ are large. Thus, decision rules can be established by linking them to the cost thresholds $h_1$ and $h_2$, whereby, for example, adoption decision is made whenever $r_1 < h_1$.

As indicated earlier, for some situations, only Phase I learning is necessary and Phase II is not required. However, in the case of our aviation scenario, as indicated in the previous section, both Phase I and Phase II are necessary, with Rule I used for Phase I, and Rule IIR used for Phase II. In this case, assuming we have learned the usefulness of the given policy, say, to put the particular aircraft in service, Rule IIR would instead look for punishments that may cause termination of the service to safeguard the safety and well-being of its passengers. We note that Rule IIR represents a stricter criterion with a stronger propensity to termination, since discontinuation would be harder if one uses Rule IR instead: we may decide to discontinue the service if there is an accumulation of $m'$ punishments, not necessarily consecutive. From (11), the mean number of observations E[$Z'$] relating to Rule IIR for Phase II is, by the Reflection Principle,

$$\mathrm{E}[\boldsymbol{Z}'] = \frac{m'}{q}. \tag{15}$$

Similarly, from (5), the mean number of observations E[$X'$] relating to Rule IR for Phase II is, again by the Reflection Principle,

$$\mathrm{E}[\boldsymbol{X}'] = \frac{1-q^m}{pq^m}. \tag{16}$$

The associated cost ratios are summarized in Table 2 below.

Simulation experiments are performed to gauge the accuracy of the above results. These are shown in Table 3 below. It compares the average values and the standard deviations, with the latter corresponding to the square roots of the variances determined above. For a given combination of parameters, 100,000 trial episodes are performed; the expected values and standard deviations are calculated based on these 100,000 episodes. The error percentages are computed as follows:

**Table 2.** Cost ratios for the two phases.

|  | **PHASE I** | **PHASE II** |
|---|---|---|
| **MISSION CRITICAL SYSTEMS** | $(1 - p^m) / (mqp^m)$ | $1/q$ |
| **INTERMEDIATE LEVEL 2** | $(1 - p^m) / (mqp^m)$ | $(1 - q^{m'}) / (m'pq^{m'})$ |
| **INTERMEDIATE LEVEL 1** | $1/p$ | $1/q$ |
| **NON-MISSION CRITICAL SYSTEMS** | $1/p$ | $(1 - q^{m'}) / (m'pq^{m'})$ |

$$Err = \frac{|\text{theoretical prediction} - \text{empirical measurement}|}{\text{empirical measurements}} \times 100\%.$$

**Table 3.** Comparison of the average trialing costs and the standard deviations.

| m | p | E[X] (th) | E[X] (expt) | Err (%) | E[Z] (th) | E[Z] (expt) | Err (%) | std. [X] (th) | std. [X] (expt) | Err (%) | std. [Z] (th) | std. [Z] (expt) | Err (%) |
|---|---|---|---|---|---|---|---|---|---|---|---|---|---|
|   | 0.6 | 9.07 | 9.05 | 0.247 | 5.0 | 4.99 | 0.200 | 7.01 | 7.01 | 0.018 | 1.83 | 1.81 | 0.863 |
| 3 | 0.75 | 5.48 | 5.47 | 0.018 | 4.0 | 4.00 | 0.000 | 3.40 | 3.38 | 0.307 | 1.15 | 1.15 | 0.006 |
|   | 0.9 | 3.71 | 3.72 | 0.129 | 3.33 | 3.33 | 0.100 | 1.46 | 1.46 | 0.584 | 0.61 | 0.61 | 0.355 |
|   | 0.6 | 29.65 | 29.77 | 0.397 | 8.33 | 8.35 | 0.199 | 26.00 | 26.16 | 0.599 | 2.36 | 2.36 | 0.142 |
| 5 | 0.75 | 12.86 | 12.84 | 0.121 | 6.67 | 6.66 | 0.100 | 9.31 | 9.31 | 0.035 | 1.49 | 1.49 | 0.142 |
|   | 0.9 | 6.94 | 6.93 | 0.031 | 5.56 | 5.56 | 0.080 | 3.24 | 3.23 | 0.469 | 0.79 | 0.78 | 0.253 |
|   | 0.6 | 86.81 | 87.02 | 0.242 | 11.67 | 11.67 | 0.029 | 81.44 | 81.91 | 0.673 | 2.79 | 2.79 | 0.292 |
| 7 | 0.75 | 25.97 | 25.85 | 0.461 | 9.33 | 9.34 | 0.071 | 20.89 | 20.84 | 0.226 | 1.76 | 1.77 | 0.355 |
|   | 0.9 | 10.91 | 10.91 | 0.057 | 7.77 | 7.78 | 0.029 | 5.79 | 5.80 | 0.142 | 0.93 | 0.93 | 0.405 |
|   | 0.6 | 410.95 | 412.69 | 0.422 | 16.67 | 16.67 | 0.019 | 402.81 | 405.22 | 0.597 | 3.33 | 3.34 | 0.181 |
| 10 | 075 | 67.03 | 67.05 | 0.030 | 13.33 | 13.33 | 0.025 | 59.51 | 59.37 | 0.244 | 2.11 | 2.10 | 0.334 |
|   | 0.9 | 18.68 | 18.67 | 0.037 | 11.11 | 11.11 | 0.010 | 7.01 | 7.01 | 0.018 | 1.83 | 1.81 | 0.863 |

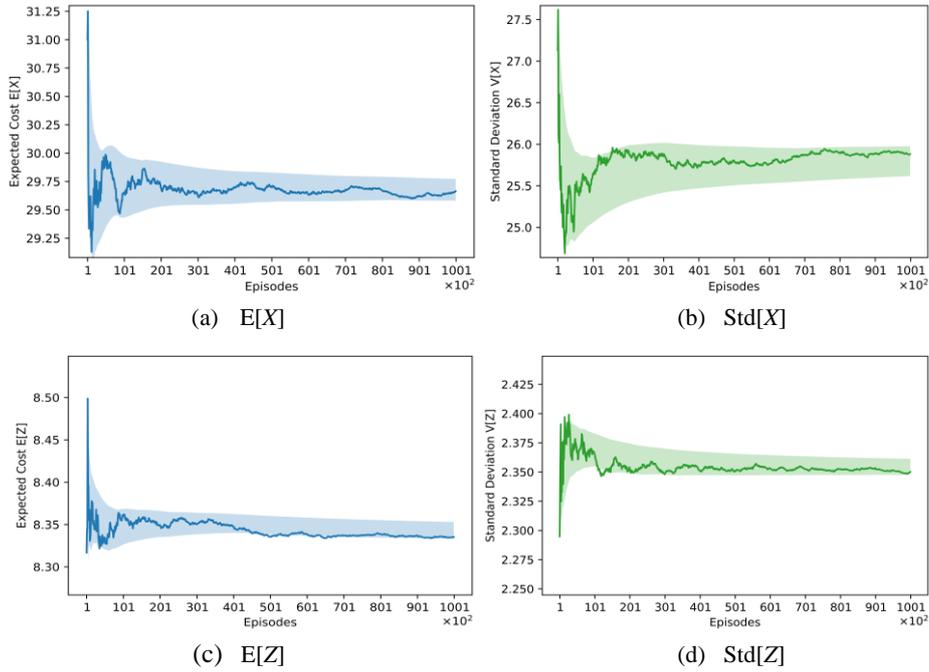

(a) E[X]  (b) Std[X]

(c) E[Z]  (d) Std[Z]

**Fig. 2.** Simulation results with $p = 0.6$, $m = 5$.

We see that the agreement is quite acceptable in all cases, with error below 1%. Figure 2, plots the experimental data for the case $p = 0.6$, $m = 5$. We see that some significant transient fluctuations are evident in the first 200 episodes, but gradually settles to an equilibrium after around 400 episodes. While some fluctuations are still present thereafter, they eventually converge to the values as predicted by the theory. Although we have not shown the behavior for other parameter settings, they behave in much the same way as those shown in Figure 2.

## 5  Summary and Conclusion

In this paper, we have studied the practical situation of learning the usefulness of a given policy for adoption by repeated sequential trials, each can result in a reward or a punishment. The entire evaluation process may be divided into two distinct phases, one for assessing initial acceptability, and one for assessing ongoing feasibility. Due to a large number of unknown factors and incomplete information, the outcome of each trial is subject to probabilistic variations and cannot be predicted exactly.

Such learning process requires suitable stopping criteria in order for the results of the observations to be consolidated and learned. Here, the probabilistic influence of the learning environment is explicitly modeled, where the outcome of each observational trial is taken to be independent and identically distributed. Four operational stopping rules, applicable to varying levels of mission-critical requirements, are established that are applicable to the two phases of learning reinforcement.

The performance of these rules are analyzed, and closed-form expressions of key measures of interest are given. In particular, cost ratios are obtained for the two phases of learning for system operations exhibiting different characteristics, and decision rules linking the trial outcomes to policy choices are developed. Experimentations have also been carried out, and the experimental results exhibit good agreements with the theoretical findings.

The present study is applicable to a wide variety of learning situations in an unknown environment based on rewards and punishments. The proposed method is useful in helping to arrive at sound operational decisions, and the associated costs of systematic evaluation has been calculated. While here we have adopted an independent, identically distributed set of random variables for the outcomes, future studies may relax on this assumption and examine situations where the outcomes are Markov dependent or where the underlying random variables are not identically distributed; doing so should be able to further enhance the usefulness of these results.